\documentclass[runningheads]{llncs}

\usepackage{graphicx}

\usepackage{hyperref}

\usepackage{booktabs}
\usepackage{multirow}
\usepackage[utf8]{inputenc}
\usepackage{subcaption}
\usepackage{multirow}
\usepackage{url}
\usepackage{makecell}
\usepackage[colorinlistoftodos]{todonotes}
\usepackage{subcaption}

\newcommand{\specialcell}[2][c]{%
  \begin{tabular}[#1]{@{}c@{}}#2\end{tabular}}

\begin{document}

\title{Fake news as we feel it: perception and conceptualization of the term\\``fake news'' in the media}
\titlerunning{Perception and conceptualization of the term ``fake news'' in the media}

\author{Evandro Cunha\inst{1,2}
\and
Gabriel Magno\inst{1}
\and
Josemar Caetano\inst{1}
\and
\\Douglas Teixeira\inst{1}
\and
Virgilio Almeida\inst{1,3}
}

\authorrunning{E. Cunha et al.}

\institute{Dept. of Computer Science, Universidade Federal de Minas Gerais (UFMG), Brazil
\email{\{evandrocunha, magno, josemarcaetano, douglas, virgilio\}@dcc.ufmg.br}\\
\and
Leiden University Centre for Linguistics (LUCL), The Netherlands
\and
Berkman Klein Center for Internet \& Society, Harvard University, USA
}

\maketitle

\begin{abstract}
In this article, we quantitatively analyze how the term ``fake news'' is being shaped in news media in recent years. We study the perception and the conceptualization of this term in the traditional media using eight years of data collected from news outlets based in 20 countries. Our results not only corroborate previous indications of a high increase in the usage of the expression ``fake news'', but also show contextual changes around this expression after the United States presidential election of 2016. Among other results, we found changes in the related vocabulary, in the mentioned entities, in the surrounding topics and in the contextual polarity around the term ``fake news'', suggesting that this expression underwent a change in perception and conceptualization after 2016. These outcomes expand the understandings on the usage of the term ``fake news'', helping to comprehend and more accurately characterize this relevant social phenomenon linked to misinformation and manipulation.

\end{abstract}

\section{Introduction}
\label{sec:intro}

The term ``fake news'', defined as ``false, often sensational, information disseminated under the guise of news reporting''~\cite{collins-woty}, gained so much attention that it was named the Collins Word of the Year 2017 due to its unprecedented usage increase of 365\% in the Collins Corpus~\cite{collins-woty}.
Even though the concept of news articles aimed to mislead readers is by no means new~\cite{fake-history}, it seems to exist a relationship between the very expression ``fake news'' with the 2016 presidential election in the United States of America: Davies \cite{now-woty}, using data from the NOW Corpus, shows that ``there is almost no mention of `fake news' until the first week of November [2016] (...) and then it explodes in Nov 11-20, and has stayed very high since then''. The author adds that the reason ``why people all of the sudden started talking about something that had really not been mentioned much at all until that time'' was ``the US elections, which were held on November 9, 2016''.

The sudden popularization of an already existing term (that is, not a neologism) in a language poses interesting questions regarding how concepts around this term are perceived by the speakers of that language. We might ask, for instance: what changed (if anything) in terms of conceptualization of this expression after its boom? Was there any kind of shift in the meaning of this expression when it became widely employed? If so, was this shift uniform across different varieties of the language? 
These are some of the issues of interest in \textit{lexicology}, the area of linguistics focused in the study of the lexicon, that has been fostered thanks to advances in \textit{corpus linguistics}, concerned with the use of big real-world corpora to the study of language.

The goal of this article is to provide a closer look at how newspapers and magazines across the world shape the term ``fake news'' -- which is a relevant social phenomenon linked to misinformation and manipulation, and that has been facilitated by the rise of the Internet and online social media in recent years. We investigate the perception and the conceptualization of this expression through the quantitative analysis of a large corpus of news published in 20 countries from 2010 to 2018, thus making it possible to examine not only the diachronic development of this term, but also its synchronic usage in different parts of the English-speaking world. We complement our investigation with data collected from online search queries that help to measure how the public interest in the expression ``fake news'' and in the concepts around it changed over time in different places.

\subsection{Related work}
\label{sec:related}

In 2010, Michel et al.~\cite{michel2011quantitative} coined the term~\textit{culturomics} meaning a method to study human behavior, cultural trends and language change through the quantitative diachronic analysis of texts, including of digitized books provided by the project Google Books.
Several studies explore this method to investigate topics such as the dynamics of birth and death of words~\cite{petersen2012statistical}, semantic change~\cite{gulordava2011distributional}, emotions in literary texts~\cite{acerbi2013expression} and general characteristics of modern societies~\cite{roth2014fashionable}.
However, many criticisms arose regarding limitations of inferences derived from the analysis of Google Books due to factors that range from optical character recognition errors and overabundance of scientific literature~\cite{pechenick2015characterizing} to the lack of metadata in the corpus~\cite{koplenig17}.

Leetaru~\cite{leetaru11} proposes a somewhat complementary approach that he calls~\textit{culturomics 2.0}, which uses historical news data instead of books and can, according to the author, ``yield intriguing new understandings of human society''.
In the same vein, Flaounas et al. analyze the European mediasphere~\cite{flaounas2010structure} and the writing style, gender bias and the popularity of particular topics~\cite{flaounas13} in large corpora of news articles.
Landsall-Welfare et al.~\cite{lansdall2014coverage}, also using a large dataset of media reports, observe a change of framing and sentiment associated with nuclear power after the Fukushima nuclear disaster.
They detected effects on attention, sentiment, conceptual associations and in the network of actors and actions linked to nuclear power following the accident.

In this investigation, we combine many of the methods employed in the
related works mentioned above.
However, as far as we are concerned, this is the first paper that uses these methods to examine in details how the relevant term ``fake news'' 
is being reported by news media in different parts of the world and in two distinct periods in the history of this expression. 

\subsection{Research question}
Our main research question is: \textit{was the rise of the public interest in the term ``fake news'' accompanied by changes in its conceptualization and in the perception about it?}
Based on sociolexicological theories, that  defend the existence of a considerable relationship between linguistic and extralinguistic factors with regards to the vocabulary of a language~\cite{matore53,cambraia2013lexicologia}, our hypothesis is that the change of interest in the phenomenon \textit{fake news} might have altered the general usage of the expression referring to it.
Indeed, the results obtained in our investigations indicate, in general, a positive answer for our research question. Among other findings, we show modifications in the related vocabulary and in the mentioned entities accompanying the term ``fake news'', in addition to changes in the topics associated to this concept and in the overall contextual polarity of the pieces of text around this expression in media articles after 2016.

\vspace{1em}

This article is structured as follows: in the next section, 
we present the process of acquisition and preparation of the main data source used in our investigations; in Section \ref{sec:analyses}, we describe our analyses, present the found results and discuss their implications; finally, in Section \ref{sec:conclusions}, we summarize the outcomes of our study and conclude this paper by discussing possible future outlooks.




\section{Data source}
\label{sec:data}

The main dataset used in this study comes from the Corpus of News on the Web (NOW Corpus), which contains articles from online newspapers and magazines written in English and based in 20 different countries from 2010 to the present time~\cite{davies13now}.
This corpus is available for download and online exploration at \url{https://corpus.byu.edu/now/} and, according to its authors, ``is [as of April 2018] by far the largest corpus (of any language) that is available in full-text format''.
Our analyses are relative to a version of the corpus available in the month of April 2018, containing around six billion words of data.

In this dataset, we searched for all the occurrences of the term ``fake news''. For each occurrence, the online version of the corpus provides a concordance line, or \textit{context} -- that is, a piece of text of approximately 20-30 words around (before and after) the searched term. For example, for a certain news article published in July 25 2017 in the Kenyan newspaper Daily Nation, the context around the term ``fake news'' is: \textit{(...) of social media and a study that said 90 per cent of Kenyans had encountered \textbf{fake news}. WhatsApp and Facebook are the two leading sources of misinformation, often (...)}.
All of our analyses were performed in these contexts, since words immediately surrounding a key term are more relevant to the conceptualization of this term than words further away from it, though in the same text. Wynne~\cite{wynne2008searching} adds that the main reason for using keywords in context (KWICs) in corpus linguistics is that ``interesting insights into the structure and usage of a language can be obtained by looking at words in real texts and seeing what patterns of lexis, grammar and meaning surround them''.

The total number of occurrences of ``fake news'' extracted from the NOW Corpus in April 30 2018 is 41,124. These occurrences encompass news articles published in all the 20 countries represented in the corpus, that were grouped in six regions based on their geographical locations (Africa, British Isles, Indian subcontinent, Oceania, Southeast Asia and The Americas), since it has been observed that offline and online news outlets tend to give preference to local and national news, to domesticate news about other countries and to reflect imbalanced information flows between the developed and the developing worlds~\cite{berger2009internet}.

These occurrences also cover each year in the corpus (from 2010 to 2018). Due to the previously observed increase in the usage of the term ``fake news'' during and after the 2016 presidential election in the United States of America (mentioned in Section \ref{sec:intro}), we categorized the occurrences in two periods: before and after the 2016 US election. The election was held in November, but we set the delimitation date between these periods in the end of the first semester of 2016 (June 30) in order to include the political campaign in the period \textit{after US election}.
Table \ref{tab:now-corpus} shows the number of contexts containing the term ``fake news'' in our dataset according to the geographical origin of the corresponding news media and the year and period of publication of the news article.

\begin{table}[!ht]
\centering
\caption{Number of contexts containing the term ``fake news'' in our dataset according to (a) the geographical origin of the corresponding news media and (b) the year and period (before or after the 2016 presidential election in the United States of America) of publication of the news article.}
 \vspace{.5em}
 \begin{tabular}{l|l|r}
 \multicolumn{3}{@{}l}{(a) Geographical origin of news media}\\
 \toprule
 \textbf{Region} & \textbf{Country} & \textbf{Occurrences} \\
 \midrule
 \multirow{4}{*}{\parbox{1.5cm}{Southeast Asia}} & Singapore & 3,722 \\
    & Malaysia & 3,455 \\
    & Philippines & 3,058 \\
    & Hong Kong & 171 \\
 \cmidrule{1-3}
 \multicolumn{3}{r}{Total: 25,3\% / 10,406}\\
 \midrule
 \multirow{3}{*}{\parbox{1.5cm}{The Americas}} & United States & 6,775 \\
    & Canada & 2,960 \\
    & Jamaica & 124 \\
 \midrule
 \multicolumn{3}{r}{Total: 24,0\% / 9,859}\\    
 \midrule 
 \multirow{2}{*}{British Isles} & Great Britain & 4,213 \\
    & Ireland & 2,035 \\
 \midrule
 \multicolumn{3}{r}{Total: 15,2\% / 6,248}\\     
 \bottomrule
 \end{tabular}
 \quad
 \begin{tabular}{l|l|r}
 \toprule
 \textbf{Region} & \textbf{Country} & \textbf{Occurrences} \\
 \midrule
 \multirow{5}{*}{Africa} & South Africa & 2,493 \\
    & Nigeria & 1,974 \\
    & Kenya & 1,368 \\    
    & Ghana & 300 \\
    & Tanzania & 1 \\
 \midrule
 \multicolumn{3}{r}{Total: 14,9\% / 6,136}\\ 
 \midrule
 \multirow{2}{*}{Oceania} & Australia & 3,052 \\
    & New Zealand & 1,446 \\
 \midrule
 \multicolumn{3}{r}{Total: 10,9\% / 4,498}\\     
  \midrule
 \multirow{4}{*}{\parbox{1.5cm}{Indian subcontinent}} & India & 2,961 \\
    & Pakistan & 772 \\
    & Sri Lanka & 147 \\
    & Bangladesh & 97 \\    
 \midrule
 \multicolumn{3}{r}{Total: 9,7\% / 3,977}\\     
 \bottomrule
 \end{tabular}
 \vspace{1em}
 \newline
 \begin{tabular}{l|c|c|c|c|c|c|c|c|c}
 \multicolumn{10}{@{}l}{(b) Year and period of publication of news article}\\
 \toprule
 \textbf{Year} & 2010 & 2011 & 2012 & 2013 & 2014 & 2015 & 2016 & 2017 & 2018\\
 \midrule
 \textbf{Occurrences} & 24 & 43 & 57 & 64 & 89 & 95 & 4,766 & 25,293 & 10,693 \\
 \midrule
 \textbf{Period} & \multicolumn{7}{c}{before US election} & \multicolumn{2}{c}{after US election}\\
 \midrule
 \textbf{Occurrences} & \multicolumn{7}{c}{494} & \multicolumn{2}{c}{40,630}\\
 \bottomrule
 \end{tabular}
 \label{tab:now-corpus}
\end{table}

\section{Analyses and results}
\label{sec:analyses}

In this section, we display and examine the outcomes of our investigations. Each analysis is introduced by a description of how it is able to contribute answering to our research question, followed by the methodology employed and finally by a presentation and discussion of the results found.

\subsection{Web search behavior}
\label{sec:web-search}

Before analyzing the data obtained from the NOW Corpus, we investigate whether it is possible to observe a change in Web search behavior regarding the expression ``fake news'' corresponding to the high increase in its use during and after the 2016 presidential election in the United States of America mentioned in Section \ref{sec:intro}.

\begin{figure}
\centering
  \includegraphics[width=0.6\linewidth]{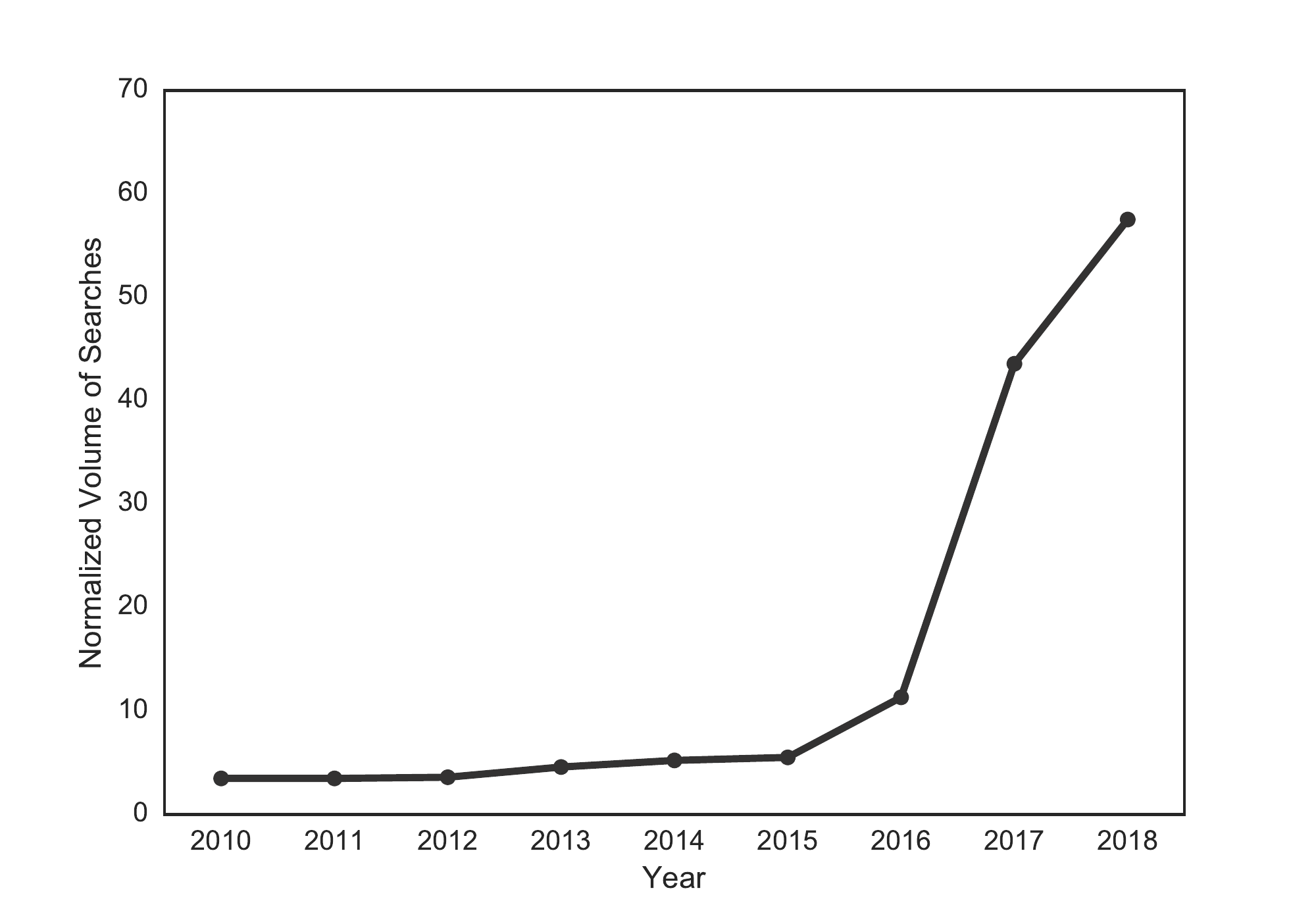}
  \caption{Normalized volume of searches for the expression ``fake news'' on Google Search from 2010 to 2018.}
  \label{fig:volume-searches-yearly}
\end{figure}

Data obtained from Google Trends\footnote{\url{https://trends.google.com/trends/}}, an online tool that indicates the frequency of particular terms in the total volume of searches in the Google Search engine, displays that public interest in the term ``fake news'' was approximately constant from 2010 until 2016, when it greatly and suddenly increased, as indicated by Figure~\ref{fig:volume-searches-yearly}.
This data also shows that, in the period before the 2016 US presidential elections, most of the countries with the highest proportions of searches for the term ``fake news'' were from the Eastern world. However, after the US election, the proportion of searches for this expression in Western countries increased considerably, especially in Europe. The 10 countries with the highest proportion of searches for the term ``fake news'' in both periods are listed in Table~\ref{tab:evolution-country}.

\begin{table}[!ht]
\centering
\caption{Countries with the highest proportion of searches for ``fake news'' on Google Search before and after US election.}
\vspace{.5em}
\label{tab:evolution-country}
\begin{tabular}{c|c}
\toprule
\textbf{Period} & \textbf{Countries} \\
\midrule
\multirow{2}{*}{before US election} & India, United Arab Emirates, Singapore, United States,\\ & Macedonia, Qatar, New Zealand, Canada, Pakistan, Australia \\
\midrule
\multirow{2}{*}{after US election} & Singapore, Philippines, United States, Canada, South Africa,\\ & Norway, Denmark, Ireland, United Kingdom, Switzerland \\
\bottomrule
\end{tabular}
\end{table}

A closer look at the data from Google Trends also reveals that the great increase in the public interest for the expression ``fake news'' coincided with a change in the focus of Web searches. Table~\ref{tab:evolution-search} shows the five most frequent search terms employed by users who also searched for ``fake news'' in the periods before and after the US election.
We observe that, before the US election, searches for ``fake news'' were generic and regarded topics related to the media industry itself, like ``article'', ``stories'' and ``report''; after the US election, however, these searches started to be more focused on political affairs and in the spread of fake news, mentioning entities like the elected president of the United States of America in 2016 (Donald Trump), the television news channel CNN (that devotes large amounts of its coverage to US politics) and the social media Facebook (considered a major source of fake news on the Internet).

\begin{table}[!ht]
\centering
\caption{Most frequent search terms related to ``fake news'' on Google Search before and after US election.}
\vspace{.5em}
\label{tab:evolution-search}
\begin{tabular}{c|c}
\toprule
\textbf{Period} & \textbf{Search terms} \\
\midrule
\multirow{2}{*}{before US election} & fake news generator, fake news article,\\ & fake news stories, make fake news, fake news report \\
\midrule
\multirow{2}{*}{after US election} & trump news, the fake news,\\ & fake news trump, cnn news, fake news facebook \\
\bottomrule
\end{tabular}
\end{table}

In this section, we used data obtained from the Google Trends tool. From the next section on, however, all of our analyses use the data described in Section \ref{sec:data}, obtained from the NOW Corpus.

\subsection{Co-occurring named entities}
\label{sec:analyses-entities}

The analysis of \textit{named entities} -- that is, real-world entities such as persons, organizations and locations that can be denoted with proper names \cite{tjong2003introduction} -- co-occurring with certain terms is an interesting way to contextualize these concepts. In our case, by identifying which entities are linked to the expression ``fake news'' in different periods of time and in different parts of the world, we are able to observe relationships of ``who and where'' in the recent history of our key-term.

In our dataset of news articles, we employed a simple method to identify named entities: we made use of the fact that newspapers and magazines consistently capitalize nouns representing named entities and counted all the words that appear capitalized in the contexts; then, we manually analyzed the most frequent capitalized words in each subdivision of the corpus (i.e. representing each region and period) to remove words not relative to named entities (such as ``I'', ``SMS'', ``March'' and words capitalized for other reasons) and to merge duplicated entities represented more than once (e.g. ``Donald'' and ``Trump'').

Table~\ref{tab:entities} shows the five most mentioned named entities in the periods before and after the 2016 US presidential election, regardless of geographical origin of the corresponding news media. Before the US election, it is possible to observe a strong connection between humor and fake news: with exception of Facebook, all the other most mentioned named entities are related to satirical TV shows and hosts based in the United States of America. On the other side, in the period after the US election, there is a movement towards politically related entities (Donald Trump), traditional media sources (CNN) and social networking services (Facebook and Twitter).
It is interesting to notice that this shift matches the already mentioned (in Table \ref{tab:evolution-search}) shift of interest towards political affairs and the spread of fake news observed in Web searches.

\begin{table}[!ht]
\centering
\caption{Most mentioned entities in the periods before and after US election.}
\vspace{.5em}
\label{tab:entities}
\begin{tabular}{c|c}
\toprule
\textbf{Period} & \textbf{Entities} \\
\midrule
\multirow{2}{*}{before US election} & The Daily Show, Jon Stewart,\\ & Onion News Network, Facebook, Stephen Colbert\\
\midrule
after US election & Donald Trump, Facebook, US, CNN, Twitter\\
\bottomrule
\end{tabular}
\end{table}

When we make this same diachronic comparison, but now considering the geographical origin of the corresponding news media, we observe a noteworthy phenomenon: the global standardization of the named entities related to \textit{fake news}. Table \ref{tab:entities-regions} indicates that local entities are more relevant in the period before the US election, when names of geographical regions (Ekiti), countries (Nigeria, China), local political parties (PDP -- People's Democratic Party of Nigeria, BJP -- Bharatiya Janata Party of India) and local personalities (Shahid Afridi, King Salman, Korina Sanchez) appear frequently among the most mentioned entities. In the contexts after the US election, however, Donald Trump, Facebook and US are the three most mentioned entities for nearly all the regions -- with the sole exception of The Americas, where CNN replaces US. 



\begin{table}[!ht]
\small
\centering
\caption{Most mentioned entities in the periods before and after US election, considering the geographical origin of the corresponding news media.}
\vspace{.5em}
 \begin{tabular}{c|c|c}
 \toprule 
  \textbf{Region}& \specialcell[c]{\textbf{Period}} & \textbf{Entities} \\
 \midrule
   \multirow{2}{*}{\specialcell[c]{Africa}} & before & \specialcell[c]{PDP, Ekiti, Nigeria}\\
  & after &\specialcell[c]{Donald Trump, Facebook, US}\\
 \midrule
  \multirow{2}{*}{\specialcell[c]{British Isles}} & before &  \specialcell[c]{Facebook, The Daily Show, Stephen Colbert} \\
  & after &\specialcell[c]{Donald Trump, Facebook, US} \\ 
 \midrule
  \multirow{2}{*}{\specialcell[c]{Indian subcontinent}} & before & \specialcell[c]{Shahid Afridi, King Salman of Saudi Arabia, BJP} \\
  & after &\specialcell[c]{Facebook, Donald Trump, US}\\
 \midrule
  \multirow{2}{*}{\specialcell[c]{Oceania}} & before & \specialcell[c]{Twitter, The Daily Show, NBC} \\
  & after &\specialcell[c]{Donald Trump, Facebook, US} \\
  \midrule
    \multirow{2}{*}{\specialcell[c]{Southeast Asia}} & before &\specialcell[c]{Korina Sanchez, US, China} \\
  & after &\specialcell[c]{Facebook, Donald Trump, US} \\
  \midrule
 \multirow{2}{*}{\specialcell[c]{The Americas}} & before &\specialcell[c]{The Daily Show, Jon Stewart, Onion News Network} \\
  & after & \specialcell[c]{Donald Trump, Facebook, CNN} \\
 \bottomrule
 \end{tabular}
 \label{tab:entities-regions}
\end{table}

\subsection{Semantic fields of the surrounding vocabulary}
\label{sec:semantic-fields}

Besides the investigation of the named entities that accompany a given key-term, the analysis of the general vocabulary co-occurring with it is also valuable. In our case, one of the possible methods of performing such analysis is by observing the semantic fields (i.e. groups to which semantically related items belong) of the words co-occurring with the expression ``fake news'' in our contexts.

For performing this task, we first lemmatized all the words in the contexts by employing the WordNet Lemmatizer function provided by the Natural Language Toolkit~\cite{bird2009natural} and using \textit{verb} as the part-of-speech argument for the lemmatization method. By applying this lemmatization, we grouped together the inflected forms of the words so that they could be analyzed as single items based on their dictionary forms (\textit{lemmas}).

Then, we used Empath~\cite{fast2016empath}, ``a tool for analyzing text across lexical categories''\footnote{\url{https://github.com/Ejhfast/empath-client}}, to classify the lemmatized words according to categories that represent different semantic fields, such as diverse topics and emotions. For every context, we calculated the percentage of words belonging to each semantic field represented by an Empath category. Due to the high number of categories predefined by Empath (194 in total), we selected eight that showed interesting results and are relevant for our discussion: \textit{government}, \textit{internet}, \textit{journalism}, \textit{leader}, \textit{negative emotion}, \textit{politics}, \textit{social media} and \textit{technology}.
By way of example, the category \textit{internet} includes 79 words such as \textit{homepage}, \textit{download} and \textit{hacker}, while the category \textit{journalism} contains 69 words, including \textit{report}, \textit{article} and \textit{newspaper}.

\begin{figure}[ht]
\centering
  \includegraphics[width=1\linewidth]{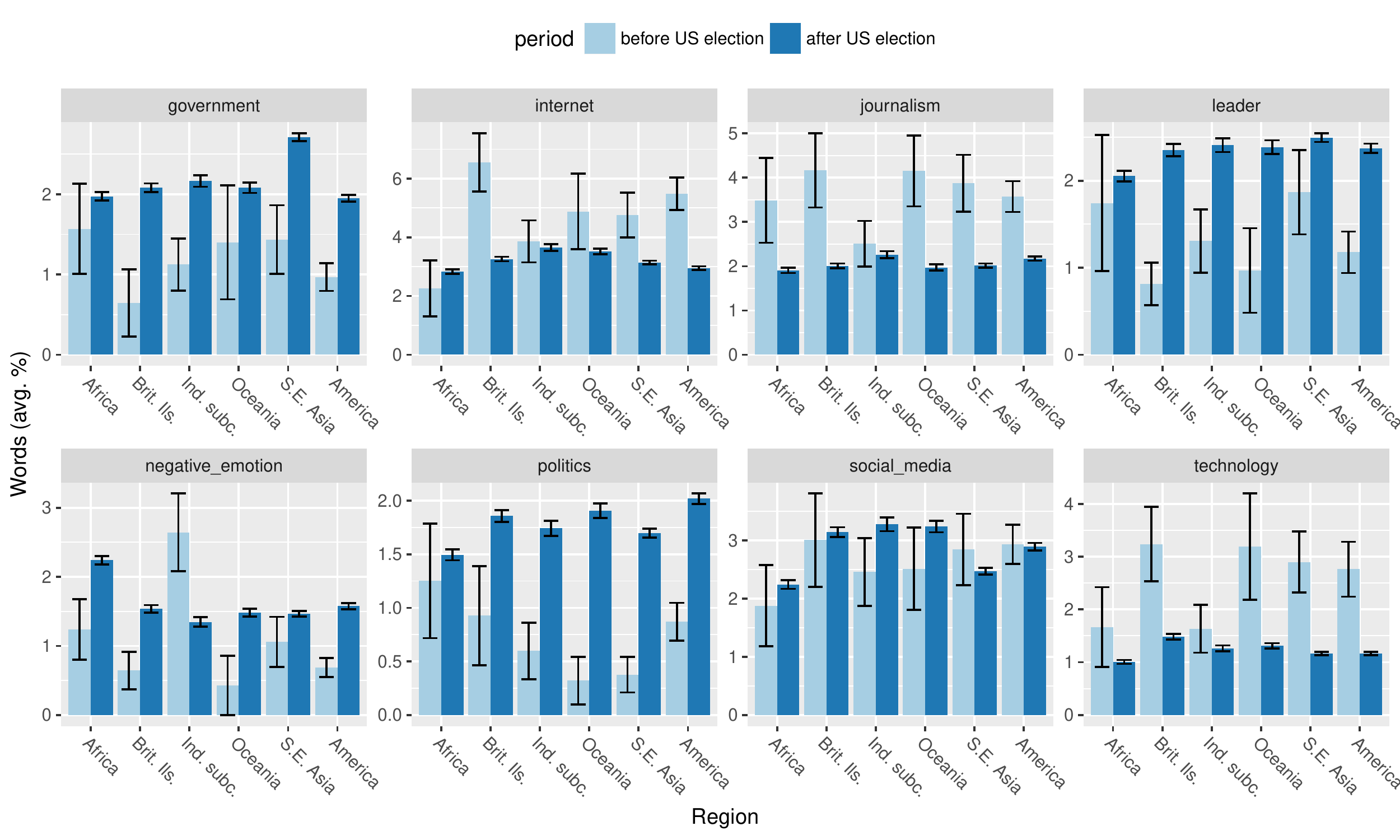}
  \caption{Percentage of words in each semantic field represented by an Empath category. Error bars indicate standard errors.}
  \label{fig:empath-avgbars}
\end{figure}

Figure~\ref{fig:empath-avgbars} displays the average percentage of words in these categories for all the six regions considered here, both before and after the 2016 US election. By analyzing the graphs presented, we observe interesting differences and trends regarding the quantitative utilization of words from the semantic fields considered. We highlight the high increase in the use of words from the related categories \textit{government}, \textit{leader} and \textit{politics} (and also from the supposedly unrelated category \textit{negative emotion}) and the high decrease in the use of words from the categories \textit{internet}, \textit{journalism} and \textit{technology} (but not \textit{social media}) in almost all regions after the US election.

We hypothesize that these results indicate a change in the focus of the news considered here: before the 2016 US election, the term ``fake news'' was probably more mentioned in contexts in which the focus was the \textit{environment} where they occur (Internet, newspapers etc.), sometimes even meta-discussions on the very topic of fake news and its dissemination; during and after the US election, however, the discussion seems to have migrated to themes more close to the \textit{content} of the fake news themselves (politics, elections etc.).



\subsection{Co-occurrence networks}
\label{sec:networks}

Another possible method of investigating the vocabulary accompanying a key-term in a corpus is through the observation of co-occurrence networks.
In our case, this method enables us to visually analyze the words that co-occur with the expression ``fake news'' in the contexts that we are considering. 
Here, 
we compare co-occurrence networks between the periods before and after the 2016 US election, regardless of the geographical origin of the media outlets. 
These networks are represented here by graphs, in which each node corresponds to a word and each edge corresponds to an association between two given words.

\begin{figure}[h!]
    \centering
    \begin{subfigure}[b]{0.49\textwidth}
        \includegraphics[width=\textwidth]{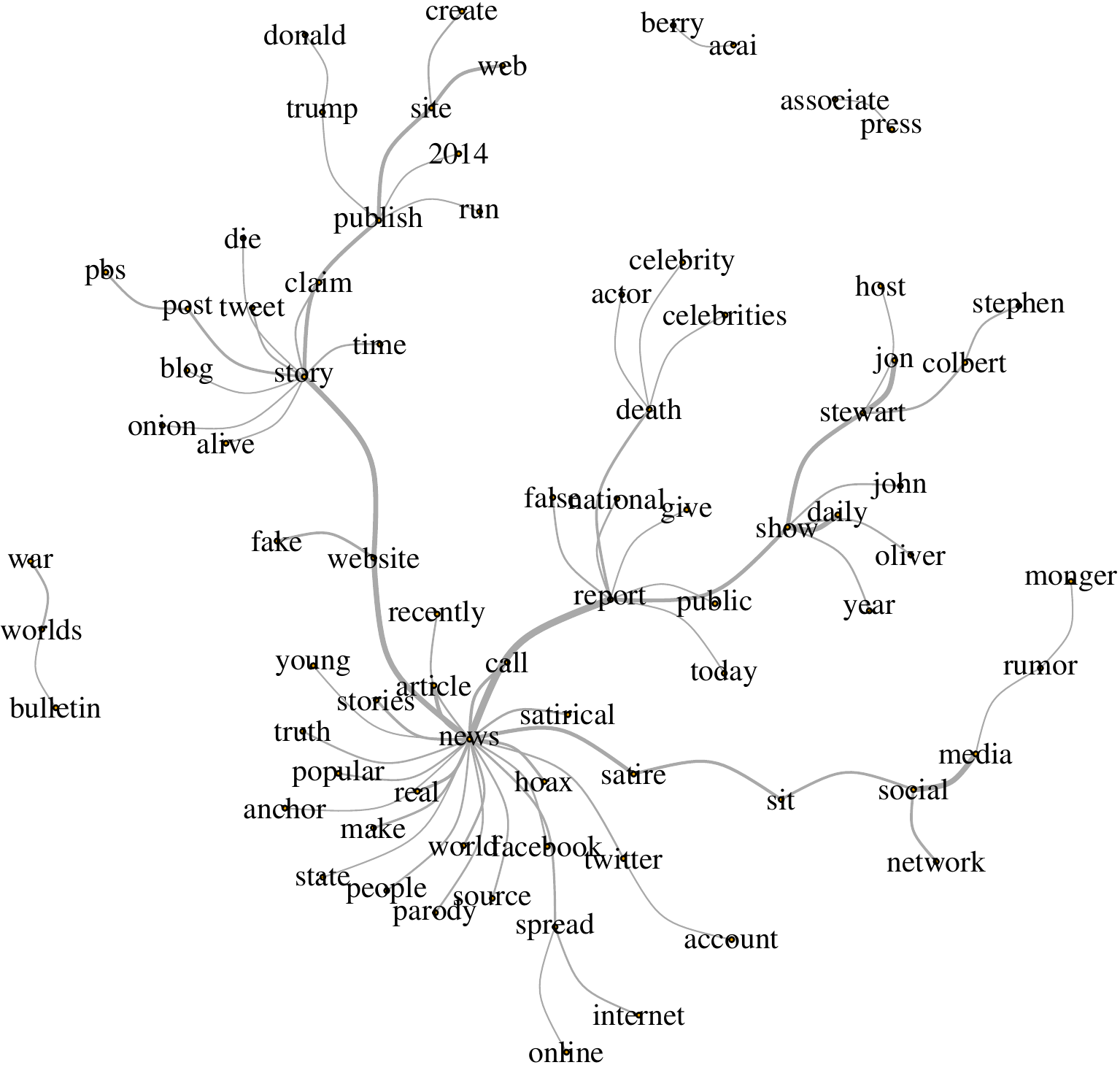}
        \caption{Before US election}
        \label{fig:cooc_net-after}
    \end{subfigure}
    \begin{subfigure}[b]{0.49\textwidth}
        \includegraphics[width=\textwidth]{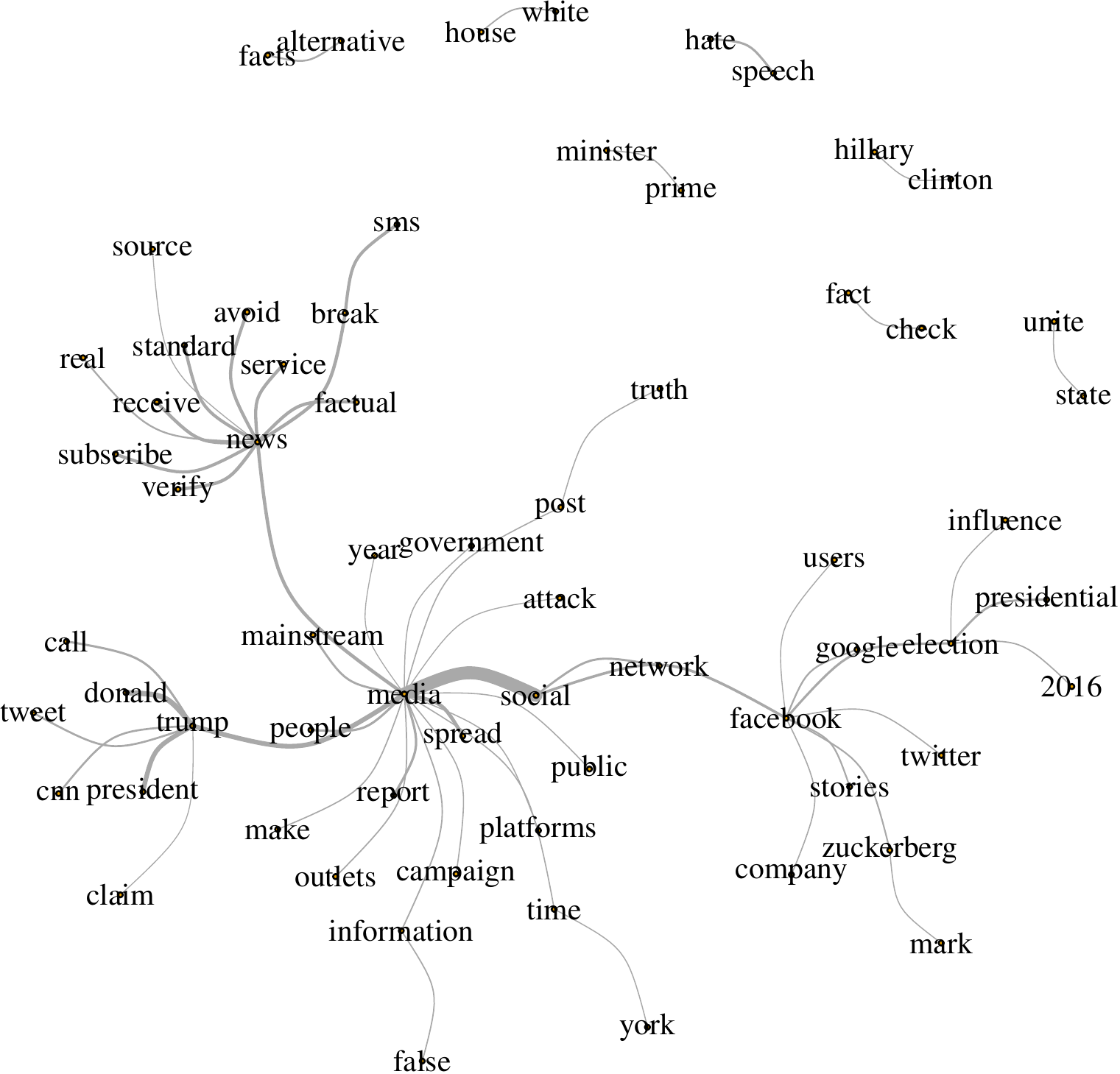}
        \caption{After US election}
        \label{fig:cooc_net-after}
    \end{subfigure}
    \caption{Co-occurrence networks of words before and after the 2016 US election.}\label{fig:cooc_net}
\end{figure}

To build our graphs of co-occurring words, we followed the steps below. 
First, we counted the number of contexts in which two given
words co-occur, so that we have the volume of co-occurrences for each pair of words.
Then, we normalized these values in order to work with percentages instead of working with the absolute number of co-occurrences.
To improve the visualization of the graphs, we filtered out minor relationships and highlighted the strongest ones by removing nodes and edges when the co-occurrence percentage was lower than $0.8$\% (for the period before the US election) and $0.5$\% (for the period after the US election), and by plotting the width of the edges proportionally to the strength of association between two given words.
Finally, we obtained the \emph{maximum spanning trees} of both graphs, which are presented in Figure \ref{fig:cooc_net}.

This method of investigation enables us to make several qualitative observations. 
Comparing the two graphs, we notice clear changes in the relationships between the words co-occurring with ``fake news'' in our contexts. For instance, before the US election, the main cluster contains words related to the news industry itself (``article'', ``stories'', ``hoax'') and to Internet (``website'', ``twitter'', ``facebook'', ``account''). Corroborating previous findings (Section \ref{sec:analyses-entities}), there is also another cluster containing words referring to satirical TV shows and hosts (``daily'', ``show'', ``colbert'', ``oliver'', ``stewart'').
In the graph representing the period after the 2016 US election, we start to observe terms linked to specific events, mainly the US election itself.
Interestingly, some terms that surround meta-discussions about fake news also appear, highlighting relevant related concepts: \textit{fact check}, \textit{hate speech}, \textit{post truth} and \textit{alternative facts}.

\subsection{Topics addressed in the contexts}
\label{sec:topics}

In addition to studying the vocabulary around a key-term, it is also possible to find the main topics addressed in the pieces of text surrounding the occurrences of the expression ``fake news'' in our corpus. 

For this task, we used latent Dirichlet allocation (LDA)~\cite{blei2003latent}, a way of automatically discovering topics discussed in texts.
First, we lowercased and tokenized all the words in the dataset. Then, we removed stop words using the list provided by the Natural Language Toolkit -- after having added the words ``fake'' and ``news'' to this list, since they appear in all contexts. Finally, we ran the LDA algorithm using \texttt{gensim}~\cite{rehurek_lrec}, a Python library for topic modeling. We used topic coherence score \cite{Newman:2010:AET:1857999.1858011} to choose the optimum number of topics \(k\) to be returned by the algorithm. Thus, for each region, we ran the LDA algorithm starting with \(k\)=2 and ending with \(k\)=20, and chose the best LDA model, that is, the LDA model with highest topic coherence score. All regions had, respectively, \(k\)=2 and \(k\)=14 for the periods before and after the US election, except The Americas, that had \(k\)=8 and \(k\)=14.
For each region, the LDA returned these \(k\) topics containing words ordered by importance in the corresponding context, filtered both by region and topic.
We then selected the most important topic 
as the representative of each region and period. Table~\ref{tab:lda} shows the main topic for each region in both periods (before and after the US election) and the top ranked ten words produced by our LDA model. 

Unlike in the period before the US election, the words related to each topic inferred by LDA are cohesive among each other in the period after the US election.
We observe, for all regions, a relevant frequency of words related to politics and social networks in the period after the US election. More specifically, the words ``russian'', ``russia'', ``election'' and ``facebook'' rank high in this period. In the period before the US election, most of the top words are related to Internet and to the spread of misinformation, in all regions.

\begin{table}[!ht]
\small
\centering
\caption{Main topic for each region. Inside each topic, ten words are presented in order of importance according to the LDA output.}
\vspace{.5em}
 \begin{tabular}{c|c|c}
 \toprule 
  \textbf{Region}& \specialcell[c]{\textbf{Period}} & \textbf{Main topic words} \\
 \midrule
   \multirow{4}{*}{\specialcell[c]{Africa}} & before & \specialcell[c]{become, world, party, south, leave,\\ week, online, state, give, member} \\
 && \\
  & after &\specialcell[c]{trump, people, spread, president, truth,\\ propaganda, thing, look, show, nigerian} \\
 \midrule
  \multirow{4}{*}{\specialcell[c]{British Isles}} & before &  \specialcell[c]{story, account, real, daily, website,\\new, show, use, death, state} \\
 && \\
  & after &\specialcell[c]{propaganda, source, russian, american, russia, \\lie, mean, popular, politics, allegation} \\ 
 \midrule
  \multirow{4}{*}{\specialcell[c]{Indian subcontinent}} & before & \specialcell[c]{create, spread, report, death, also, \\lot, say, not, social, do} \\
 && \\
  & after &\specialcell[c]{facebook, also, problem, user, issue,\\company, russian, state, work, zuckerberg} \\
 \midrule
  \multirow{4}{*}{\specialcell[c]{Oceania}} & before & \specialcell[c]{people, story, site, report, website,\\mortgage, would, fool, year, day} \\
 && \\
  & after &\specialcell[c]{election, influence, media, create, russian,\\question, policy, discuss, presidential, word} \\
  \midrule
    \multirow{4}{*}{\specialcell[c]{Southeast Asia}} & before &\specialcell[c]{article, website, story, report, site, \\celebrity, death, publish, go, viral} \\
 && \\
  & after &\specialcell[c]{public, government, fact, twitter, proliferation,\\day, official, however, phenomenon, concern\\} \\
  \midrule
 \multirow{4}{*}{\specialcell[c]{The Americas}} & before &\specialcell[c]{release, chip, firm, flurry, blue, \\ target, date, breadcrumb, irresponsible, last} \\
 && \\
  & after & \specialcell[c]{facebook, problem, network, company, also,\\believe, publish, work, policy, russian} \\
 \bottomrule
 \end{tabular}
 \label{tab:lda}
\end{table}

\subsection{Polarity}
\label{sec:polarity}

Our final analysis explores a different feature of the contexts in which the expression ``fake news'' appear in our dataset: their \textit{polarities}, that is, whether the expressed opinion in the texts is mostly positive, negative or neutral.
Here, we performed sentiment analysis~\cite{Silva:2016:SCS:2911992.2932708} in each one of the contexts in our dataset using \texttt{SentiStrength}\footnote{\url{http://sentistrength.wlv.ac.uk/}}~\cite{Thelwall2010}, a tool able to estimate the strength of positive and negative sentiment in short texts. Given a piece of text, this tool returns a score that varies from -4 (negative sentiment) to +4 (positive sentiment).

\begin{figure}
\centering
  \includegraphics[width=1\linewidth]{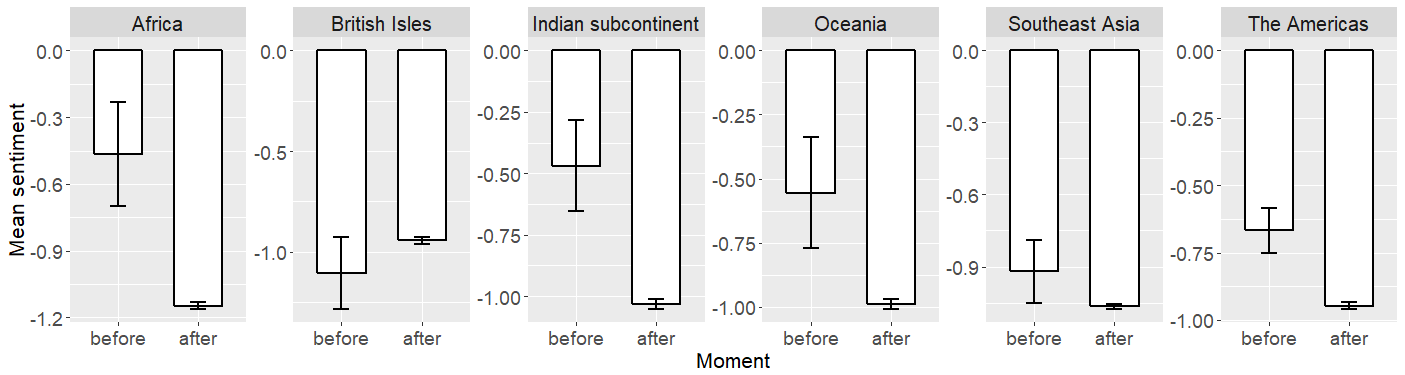}
\caption{Average polarity of the contexts in each region before and after US election (bars indicate the standard error of the mean).}
\label{fig:sentiment_period_by_region}
\end{figure}

We are interested in analyzing how polarity changes over time and in different regions when it comes to ``fake news'' and how this can be perceived in our dataset.
Figure \ref{fig:sentiment_period_by_region} depicts the average polarity of the contexts in each region before and after the 2016 US presidential election. We first observe a clear dominance of negative polarities in all periods and regions, 
indicating that the term ``fake news'' is often related to negative words~\cite{10.1371/journal.pone.0138740} and sentiments -- which is not surprising, since the concept of fake news seems to be strongly associated with negative concepts, like misinformation, manipulation and the spread of untrue facts.

In this figure, we also observe that, in general, the polarity expressed in the contexts in the period after the US election is more negative than before. The only exception is in the British Isles, where the difference of polarity between the periods is not statistically relevant. This result seems to corroborate findings presented in previous sections of this article, which demonstrated that, before the 2016 US election, the term ``fake news'' was often linked to satirical TV shows and more general topics, while in the period during and after the election the topics became more related to the spread of false information in the context of political activity.

\subsection{Summary of results}

The most relevant outcomes of the analyses presented in this section can be summarized and integrated as follows:
\begin{itemize}
\item the interest for the term ``fake news'' suddenly increased after the 2016 US election, as indicated by the rise of news about it and of Google Search queries for this expression (Section \ref{sec:web-search});
\item this growth was accompanied by a change of framing around the term ``fake news'' -- from, for instance, topics regarding the media industry itself to those related to political affairs (Sections \ref{sec:web-search}, \ref{sec:semantic-fields}, \ref{sec:networks}, \ref{sec:topics});
\item the named entities linked to the expression ``fake news'' not only changed towards political topics, but also suffered from global standardization after the US election (Section \ref{sec:analyses-entities});
\item the negativity of the news containing the term ``fake news'' increased after the US election (Section \ref{sec:polarity}).
\end{itemize}

All these results suggest that, as hypothesized in Section \ref{sec:intro}, the rise of public interest in the term ``fake news'' came with changes in its conceptualization and in the perception about it. 







\section{Concluding remarks}
\label{sec:conclusions}

Due to the increased role of the Internet in modern societies, topics regarding misinformation and manipulation in online environments seem to be subject to progressively more public debate and interest, including from the traditional media. Understanding how these topics are viewed through the eyes of opinion leaders is crucial to comprehend how public opinion about them is being shaped in present day.

In this article, we present a quantitative analysis on the perception and conceptualization of the term ``fake news'' in a corpus of news articles published from 2010 to 2018 in 20 countries. We investigate how media sources have been reporting topics related to
fake news and whether the rise of the public interest in this very expression during and after the 2016 presidential election in the United States of America was accompanied by changes of perception and shifts in sentiment about it. We observed changes in the vocabulary and in the mentioned entities around the term ``fake news'' in our corpus, in the topics related to this concept and in the polarity of the texts around it after 2016, as well as in Web search behavior of Google Search users interested in this concept.

We are also interested in understanding whether the term ``fake news'' is framed differently across the globe
-- and, if so, which are these differences. 
The existence of such variations may result in different shifts in the meanings and in the sentiments around these concepts in various regions of the world, which justifies this study as a way to more clearly understand how the public opinion is being steered in the current context in different countries of the English-speaking world.


In this paper, we analyzed the usage of the term ``fake news'' in a diachronic perspective, but only considered two historical moments: before and after a key event in the history of this expression (the 2016 US presidential election). In the future, we plan to consider a larger spectrum of periods, particularly to understand whether (and, if it is the case, when) the conceptualization of ``fake news'' changed once again. We also intend to add analyses using data from other relevant sources, including Twitter posts and Wikipedia edits.

\bibliographystyle{splncs04}
\bibliography{ref}

\begin{thebibliography}{10}
\providecommand{\url}[1]{\texttt{#1}}
\providecommand{\urlprefix}{URL }
\providecommand{\doi}[1]{https://doi.org/#1}

\bibitem{acerbi2013expression}
Acerbi, A., Lampos, V., Garnett, P., Bentley, R.A.: The expression of emotions
  in 20th century books. PLOS ONE  \textbf{8}(3),  e59030 (2013)

\bibitem{berger2009internet}
Berger, G.: How the {I}nternet impacts on international news: Exploring
  paradoxes of the most global medium in a time of `hyperlocalism'.
  International Communication Gazette  \textbf{71}(5),  355--371 (2009)

\bibitem{bird2009natural}
Bird, S., Loper, E., Klein, E.: Natural language processing with {P}ython.
  O'Reilly Media Inc. (2009)

\bibitem{blei2003latent}
Blei, D.M., Ng, A.Y., Jordan, M.I.: Latent {D}irichlet allocation. Journal of
  Machine Learning Research  \textbf{3}(Jan),  993--1022 (2003)

\bibitem{cambraia2013lexicologia}
Cambraia, C.N.: Da lexicologia social a uma lexicologia
  s{\'o}cio-hist{\'o}rica: caminhos poss{\'i}veis. Revista de Estudos da
  Linguagem  \textbf{21}(1),  157--188 (2013)

\bibitem{collins-woty}
{Collins Dictionary}: Word of the year 2017. Retrieved from
  \url{https://www.collinsdictionary.com/woty}. Accessed on May 4, 2018 (2017)

\bibitem{davies13now}
Davies, M.: Corpus of {N}ews on the {W}eb ({NOW}): 3+ billion words from 20
  countries, updated every day. Available online at
  \url{https://corpus.byu.edu/now/} (2013)

\bibitem{now-woty}
Davies, M.: Fake news. Retrieved from
  \url{https://corpus.byu.edu/now/help/fake-news.asp}. Accessed on May 4, 2018
  (2017)

\bibitem{fast2016empath}
Fast, E., Chen, B., Bernstein, M.S.: Empath: Understanding topic signals in
  large-scale text. In: Proceedings of the 2016 CHI Conference on Human Factors
  in Computing Systems. pp. 4647--4657. ACM (2016)

\bibitem{flaounas13}
Flaounas, I., Ali, O., Lansdall-Welfare, T., Bie, T.D., Mosdell, N., Lewis, J.,
  Cristianini, N.: Research methods in the age of digital journalism:
  Massive-scale automated analysis of news-content -- topics, style and gender.
  Digital Journalism  \textbf{1}(1),  102--116 (2013).
  \doi{10.1080/21670811.2012.714928}

\bibitem{flaounas2010structure}
Flaounas, I., Turchi, M., Ali, O., Fyson, N., De~Bie, T., Mosdell, N., Lewis,
  J., Cristianini, N.: The structure of the {EU} mediasphere. PLOS ONE
  \textbf{5}(12),  e14243 (2010)

\bibitem{gulordava2011distributional}
Gulordava, K., Baroni, M.: A distributional similarity approach to the
  detection of semantic change in the {G}oogle {B}ooks {N}gram corpus. In:
  Proceedings of the GEMS 2011 Workshop on GEometrical Models of Natural
  Language Semantics. pp. 67--71. Association for Computational Linguistics
  (2011)

\bibitem{koplenig17}
Koplenig, A.: The impact of lacking metadata for the measurement of cultural
  and linguistic change using the {G}oogle {N}gram data sets -- reconstructing
  the composition of the {G}erman corpus in times of {WWII}. Digital
  Scholarship in the Humanities  \textbf{32}(1),  169--188 (2017).
  \doi{10.1093/llc/fqv037}, \url{+ http://dx.doi.org/10.1093/llc/fqv037}

\bibitem{lansdall2014coverage}
Lansdall-Welfare, T., Sudhahar, S., Veltri, G.A., Cristianini, N.: On the
  coverage of science in the media: A big data study on the impact of the
  {F}ukushima disaster. In: 2014 IEEE International Conference on Big Data. pp.
  60--66. IEEE (2014)

\bibitem{leetaru11}
Leetaru, K.: Culturomics 2.0: Forecasting large-scale human behavior using
  global news media tone in time and space. First Monday  \textbf{16}(9) (2011)

\bibitem{matore53}
Matoré, G.: La méthode en lexicologie: domaine français. Didier, Paris
  (1953)

\bibitem{michel2011quantitative}
Michel, J.B., Shen, Y.K., Aiden, A.P., Veres, A., Gray, M.K., Pickett, J.P.,
  Hoiberg, D., Clancy, D., Norvig, P., Orwant, J., et~al.: Quantitative
  analysis of culture using millions of digitized books. Science
  \textbf{331}(6014),  176--182 (2011)

\bibitem{Newman:2010:AET:1857999.1858011}
Newman, D., Lau, J.H., Grieser, K., Baldwin, T.: Automatic evaluation of topic
  coherence. In: Human Language Technologies: The 2010 Annual Conference of the
  North American Chapter of the Association for Computational Linguistics. pp.
  100--108. HLT '10, Association for Computational Linguistics, Stroudsburg,
  PA, USA (2010), \url{http://dl.acm.org/citation.cfm?id=1857999.1858011}

\bibitem{pechenick2015characterizing}
Pechenick, E.A., Danforth, C.M., Dodds, P.S.: Characterizing the {G}oogle
  {B}ooks corpus: Strong limits to inferences of socio-cultural and linguistic
  evolution. PLOS ONE  \textbf{10}(10),  e0137041 (2015)

\bibitem{petersen2012statistical}
Petersen, A.M., Tenenbaum, J., Havlin, S., Stanley, H.E.: Statistical laws
  governing fluctuations in word use from word birth to word death. Scientific
  Reports  \textbf{2} (2012)

\bibitem{rehurek_lrec}
{\v R}eh{\r u}{\v r}ek, R., Sojka, P.: {Software Framework for Topic Modelling
  with Large Corpora}. In: {Proceedings of the LREC 2010 Workshop on New
  Challenges for NLP Frameworks}. pp. 45--50. ELRA, Valletta, Malta (May 2010),
  \url{http://is.muni.cz/publication/884893/en}

\bibitem{roth2014fashionable}
Roth, S.: Fashionable functions: A {G}oogle {N}gram view of trends in
  functional differentiation (1800-2000). International Journal of Technology
  and Human Interaction  \textbf{10}(2),  34--58 (2014)

\bibitem{Silva:2016:SCS:2911992.2932708}
Silva, N.F.F.D., Coletta, L.F.S., Hruschka, E.R.: A survey and comparative
  study of tweet sentiment analysis via semi-supervised learning. ACM Comput.
  Surv.  \textbf{49}(1),  15:1--15:26 (Jun 2016). \doi{10.1145/2932708},
  \url{http://doi.acm.org/10.1145/2932708}

\bibitem{fake-history}
Standage, T.: The true history of fake news. 1843 Magazine. Retrieved from
  \url{https://bit.ly/2sh9OYQ}. Accessed on May 4, 2018 (2017)

\bibitem{Thelwall2010}
Thelwall, M., Buckley, K., Paltoglou, G., Cai, D., Kappas, A.: Sentiment in
  short strength detection informal text. J. Am. Soc. Inf. Sci. Technol.
  \textbf{61}(12),  2544--2558 (Dec 2010). \doi{10.1002/asi.v61:12},
  \url{http://dx.doi.org/10.1002/asi.v61:12}

\bibitem{tjong2003introduction}
Tjong Kim~Sang, E.F., De~Meulder, F.: Introduction to the {CoNLL-2003} shared
  task: language-independent named entity recognition. In: Proceedings of the
  7th Conference on Natural Language Learning at HLT-NAACL 2003. Association
  for Computational Linguistics (2003)

\bibitem{wynne2008searching}
Wynne, M.: Searching and concordancing. Corpus linguistics. An international
  handbook  \textbf{1},  706--737 (2008)

\bibitem{10.1371/journal.pone.0138740}
Zollo, F., Novak, P.K., Del~Vicario, M., Bessi, A., Mozetič, I., Scala, A.,
  Caldarelli, G., Quattrociocchi, W.: Emotional dynamics in the age of
  misinformation. PLOS ONE  \textbf{10}(9),  1--22 (09 2015).
  \doi{10.1371/journal.pone.0138740},
  \url{https://doi.org/10.1371/journal.pone.0138740}

\end{thebibliography}

\end{document}